\documentclass{article}

\usepackage{arxiv}

\usepackage[T1]{fontenc}    
\usepackage{hyperref}       
\usepackage{url}            
\usepackage{booktabs}       
\usepackage{amsfonts}       
\usepackage{nicefrac}       
\usepackage{microtype}      
\usepackage{lipsum}
\usepackage{graphicx}
\usepackage{amsmath}        
\usepackage{amssymb}        
\usepackage[numbers]{natbib} 
\usepackage{xcolor}
\usepackage[normalem]{ulem}

\usepackage{siunitx}
\usepackage{tabularx}
\usepackage{ragged2e}
\newcolumntype{Y}{>{\RaggedRight\arraybackslash}X} 

\newcommand{\beginsupplement}{%
  \setcounter{table}{0}%
  \renewcommand{\thetable}{S\arabic{table}}%
  \setcounter{figure}{0}%
  \renewcommand{\thefigure}{S\arabic{figure}}%
  \newcounter{note}
  \renewcommand{\thenote}{S\arabic{note}}
  \newcounter{approach}
  \renewcommand{\theapproach}{S\arabic{approach}}
  \newcounter{method}
  \renewcommand{\themethod}{S\arabic{method}}
}

\title{DiSPA: Differential Substructure-Pathway Attention for Drug Response Prediction}

\author{
  Yewon Han \\
  Division of AI Software Convergence\\
  Dongguk University\\
  Seoul, South Korea \\
  \And
  Sunghyun Kim \\
  Division of Al Convergence\\
  Dongguk University\\
  Seoul, South Korea \\
  \And
  Eunyi Jeong \\
  Division of Al Convergence\\
  Dongguk University\\
  Seoul, South Korea \\
  \And
  Sungkyung Lee \\
  Department of Computer Science and AI\\
  Dongguk University\\
  Seoul, South Korea \\
  \And
  Seokwoo Yun \\
  AI Research Team\\
  Ar--ge Inc.\\
  Seoul, South Korea \\
  \And
  Sangsoo Lim\thanks{Corresponding author: \texttt{sslim@dgu.ac.kr}} \\
  Department of Computer Science and AI\\
  Dongguk University\\
  Seoul, South Korea \\
}

\begin{document}
\maketitle

\begin{abstract}
\textbf{Motivation:} Accurate prediction of drug response in precision medicine requires models that capture how specific chemical substructures interact with cellular pathway states. However, most existing deep learning approaches treat chemical and transcriptomic modalities independently or combine them only at late stages, limiting their ability to model fine-grained, context-dependent mechanisms of drug action. In addition, vanilla attention mechanisms are often sensitive to noise and sparsity in high-dimensional biological networks, hindering both generalization and interpretability. \\
\textbf{Results:} We present DiSPA (Differential Substructure-Pathway Attention), a framework that models bidirectional interactions between chemical substructures and pathway-level gene expression. DiSPA introduces differential cross-attention to suppress spurious associations while enhancing context-relevant interactions. On the GDSC benchmark, DiSPA achieves state-of-the-art performance, with strong improvements in the disjoint setting. These gains are consistent across random and drug-blind splits, suggesting improved robustness. Analyses of attention patterns indicate more selective and concentrated interactions compared to standard cross-attention. Exploratory evaluation shows that differential attention better prioritizes predefined target-related pathways, although this does not constitute mechanistic validation. DiSPA also shows promising generalization on external datasets (CTRP) and cross-dataset settings, although further validation is needed. It further enables zero-shot application to spatial transcriptomics, providing exploratory insights into region-specific drug sensitivity patterns without ground-truth validation. \\
\par\vspace{1em}
\noindent\textbf{Availability:} Source code and supplementary materials are available at \url{https://github.com/sslim-aidrug/DiSPA}. \\
\textbf{Contact:} \href{mailto:sslim@dgu.ac.kr}{sslim@dgu.ac.kr}
\end{abstract}

\section{Introduction}\label{sec:intro}
Predicting drug response remains a central challenge in precision medicine, requiring models that capture how chemical structure interacts with cellular states to determine sensitivity or resistance. Although recent approaches integrate chemical graphs, transcriptomics, and curated biological resources \citep{baptista2021deep, firoozbakht2022overview}, most methods encode these modalities independently or combine them only at late stages \citep{shen2023systematic, baptista2021deep, adam2020machine, xia2022cross}. Consequently, drug representations are often treated as static, limiting the ability to model context-dependent interactions between chemical moieties and biological pathways.

Drug substructures constitute the functional units of molecular activity, where small structural modifications can lead to substantial changes in potency \citep{stumpfe2012exploring}. In parallel, genes act collectively within pathways whose coordinated activity governs cellular responses to perturbations \citep{schubert2018perturbation}. While incorporating drug-target interaction (DTI) information can improve predictive performance \citep{pak2023improved}, existing DTI databases contain only positive associations and provide limited insight into non-targets or context-specific engagement. Moreover, static interaction priors fail to capture the dynamic cellular environments defined by heterogeneous gene expression states. As a result, current models lack a principled mechanism for linking structural determinants of drug action with pathway-level biological context.

More broadly, existing drug response models fall into two dominant paradigms: approaches that treat drug representations as invariant across contexts, and models that condition biological features on drugs only at the output level. In both settings, it remains difficult to determine whether the efficacy of drug is driven primarily by conserved chemical substructures or by context-dependent pathway activation, limiting both interpretability and generalization.

To address these limitations, we introduce DiSPA (Differential Substructure-Pathway Attention), a representation learning framework that explicitly disentangles structure-driven and context-driven mechanisms of drug response through bidirectional conditioning between chemical substructures and pathway-level gene expression. The core of DiSPA is a dual-view differential cross-attention module: in one view, pathway representations attend to drug substructures to identify structure-driven effects; in the other, drug representations attend to pathway-conditioned gene expression to capture context-dependent mechanisms. Rather than reconstructing static drug-target interactions, DiSPA learns continuous pathway-substructure and drug-pathway attention patterns that reflect dynamic molecular engagement without relying on exhaustive target annotations. Across diverse evaluation settings, DiSPA demonstrates strong predictive performance and generalization, while providing mechanistic interpretability by directly linking chemical structures and biological pathways. Furthermore, representations learned from bulk RNA-seq data transfer without retraining to single-cell and spatial transcriptomic datasets, enabling zero-shot exploration of cell type- and region-specific drug sensitivities.

\section{Methods}\label{sec:methods}

\subsection{Dataset}
We integrated the preprocessed GDSC benchmark dataset \citep{shen2023systematic} with curated pathway information from the Kyoto Encyclopedia of Genes and Genomes (KEGG) \citep{kanehisa2000kegg} to enable pathway-level modeling of drug response. The GDSC dataset provides gene expression profiles for 966 cancer cell lines, along with corresponding $IC_{50}$ measurements and annotations for 282 small-molecule drugs, including drug names and SMILES representations.
KEGG pathway definitions and gene-pathway mappings were used to associate drug-target genes with their relevant biological pathways. After harmonizing gene identifiers and filtering for consistency across data sources, the final dataset comprises 965 cell lines and 270 drugs, yielding a total of 224,078 drug-cell line response samples, with 1,692 genes mapped to 94 KEGG pathways. Detailed pathway-level mapping information, including pathway IDs, KEGG pathways, gene counts, and mapped genes, is provided in Table~\ref{stab:pathway_gene_mapping}. This integration results in a biologically structured dataset that supports pathway-aware modeling of drug-cell interactions. Data sources and preprocessing details are provided in Approaches~\ref{sapp:data_availability} and ~\ref{sapp:data_preprocessing}.

\subsection{Overview of DiSPA}
\begin{figure*}
    \centering
    \includegraphics[width=170mm]{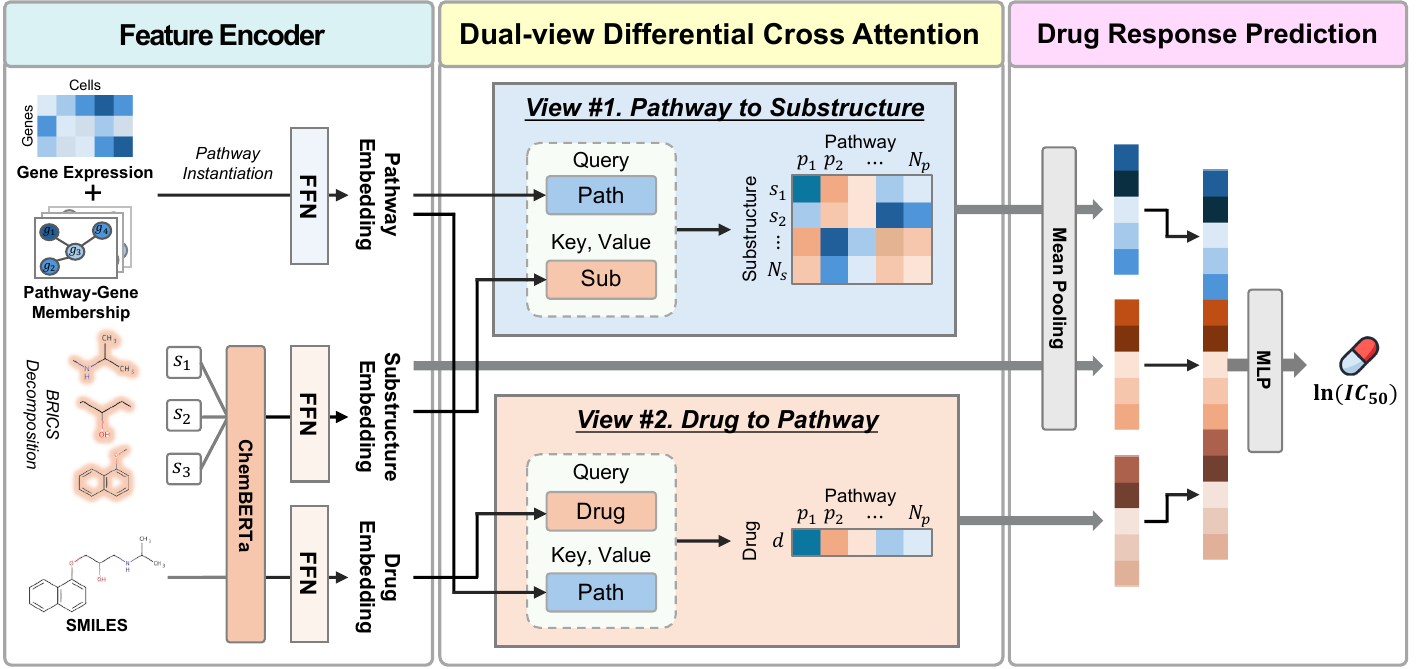}
    \caption{{\bf Overview of the DiSPA framework.} DiSPA consists of three major stages: feature encoding, dual-view differential cross-attention, and drug response prediction. Gene expression profiles are first mapped to KEGG pathways to construct pathway-level gene embeddings, while drug SMILES are decomposed into substructures and encoded alongside drug-level representations. In the dual-view cross-attention module, (View 1) pathway embeddings attend to drug substructures to capture pathway-specific chemical relevance, and (View 2) drug embeddings attend to pathway representations to model drug-conditioned biological context. The resulting pathway-substructure and drug-pathway representations are aggregated and passed through a multilayer perceptron (MLP) to predict drug response ($\ln(IC_{50})$) for each cell line-drug pair. 
    }
    \label{fig:overview}
\end{figure*}

As illustrated in Figure~\ref{fig:overview}, DiSPA is a regression framework for predicting drug response ($\ln(IC_{50})$) at the level of cell line-drug pairs, explicitly modeling interactions between chemical substructures and pathway-level gene expression.

\subsection{Data preprocessing}
\subsubsection{Gene expression embedding}
Gene expression profiles were mapped onto KEGG pathway structures to construct pathway-level representations. Expression values were first standardized across cell lines using z-score normalization.
To emphasize biologically meaningful drug response mechanisms, we restricted our analysis to KEGG pathway category 6 (Human Diseases). To verify that this choice was data-driven rather than selectively chosen, supplementary analyses showed that category 6 provided the most favorable balance of performance, efficiency, and robustness among the tested pathway sets (Tables~\ref{stab:kegg_category_performance} and~\ref{stab:kegg_full_vs_cat6}). 

For each cell line, pathway-aware gene expression was encoded as a matrix

$$
\text{{\bf E}}_{\text{path}} \in \mathbb{R}^{N_p \times N_{g}}
$$

where $N_p$ denotes the number of selected pathways and $N_g$ represents the maximum number of genes across all pathways. For a given pathway, expression values were populated only for genes belonging to that pathway, while non-member positions were zero-padded and excluded by masking. Each row of $\text{{\bf E}}_{\text{path}}$ therefore corresponds to a pathway-specific embedding that captures the expression profile of its constituent genes.

\subsubsection{Drug embedding}
Each drug was represented using its SMILES string and encoded at two complementary levels: drug-level and substructure-level. For a given drug $s$, the resulting representations are defined as
\begin{equation}
\begin{aligned}
\mathcal{E}_{\text{drug}}(s) =
\left\{
\begin{aligned}
\mathbf{E}_{\text{drug}}(s) &= \Phi(s) &&\in \mathbb{R}^{768} \\[6pt]
\mathbf{E}_{\text{sub}}(s) &= \{\Phi(s_i)\}_{i=1}^{N_s} &&\in \mathbb{R}^{N_s \times 768}
\end{aligned}
\right.
\end{aligned}
\label{eq:pre_drug}
\end{equation}

where $\Phi(\cdot)$ denotes the ChemBERTa encoder, $768$ is the dimensionality of the embedding space, and $N_s$ represents the number of substructures derived from drug $s$.
For substructure-level preprocessing, we employed the Breaking of Retrosynthetically Interesting Chemical Substructures (BRICS) algorithm \citep{degen2008art}, which decomposes molecules by cleaving chemically meaningful bonds according to predefined retrosynthetic rules. Drugs for which BRICS decomposition was not applicable-such as compounds containing metal atoms or nonstandard linkages-were excluded to ensure consistency in substructure-level representations (Table \ref{stab:brics_excluded_drugs}). This dual-level encoding enables DiSPA to capture both global molecular context at the drug level and fine-grained chemical semantics at the substructure-level, providing a rich and biologically meaningful foundation for modeling drug-pathway interactions.

\subsection{Feature encoder}
All preprocessed representations --$\mathbf{E}_{\text{path}}$, $\mathbf{E}_{\text{drug}}$, and $\mathbf{E}_{\text{sub}}$--
are transformed by modality-specific feed-forward neural networks (FFNs) that serve as feature encoders. These encoders project heterogeneous, high-dimensional inputs into a shared latent space while preserving the structural priors inherent to each modality. In particular, pathway-gene mappings encode biological organization, whereas drug- and substructure-level embeddings capture chemical semantics at different structural resolutions.

Formally, the encoded feature representations are given by

\begin{equation}
\mathbf{H}_m = \mathrm{FFN}_m(\mathbf{E}_m), 
\quad m \in \{\text{path}, \text{drug}, \text{sub}\},
\label{eq:ffn_encoder}
\end{equation}

where 
\[
\mathbf{H}_m \in 
\begin{cases}
\mathbb{R}^{N_p \times d_a}, & m = \text{path} \\
\mathbb{R}^{1 \times d_a},   & m = \text{drug} \\
\mathbb{R}^{N_s \times d_a}, & m = \text{sub}
\end{cases}
\]

Here, $\mathbf{H}_{\text{path}}$, $\mathbf{H}_{\text{drug}}$, and $\mathbf{H}_{\text{sub}}$ represent the encoded pathway-, drug-, and substructure-level features, respectively. All modality-specific FFNs project their inputs into a shared latent space of dimension $d_a$, facilitating subsequent cross-modal attention while preserving structural information unique to each modality.

\subsection{Dual-view differential cross-attention}
To model interactions between pathway-level gene features and drug substructures, we introduce a bidirectional differential cross-attention module. Unlike conventional drug response models that encode drugs and cell lines independently and fuse them only at the prediction stage, DiSPA refines each modality by explicitly conditioning it on the other during representation learning. This bidirectional design enables the model to capture context-dependent pathway-substructure associations more effectively.

The space of possible pathway-substructure interactions is large and inherently sparse, making vanilla attention mechanisms susceptible to noise and spurious correlations. To mitigate this issue, we adopt the differential attention mechanism \citep{ye2025differential}, which constructs two complementary attention components whose difference suppresses irrelevant signals and sharpens focus on informative interactions.

Formally, given query $Q$, key $K$, and value $V$, the differential attention output is defined as:
\[
A = \left( 
    \mathrm{softmax}\!\left( \frac{Q_{1} K_{1}^{\top}}{\sqrt{d}} \right) 
    - \lambda \cdot \mathrm{softmax}\!\left( \frac{Q_{2} K_{2}^{\top}}{\sqrt{d}} \right) 
\right) \tilde{V}
\]

where 
\[
[Q_{1}; Q_{2}] = Q W_{Q}, 
\quad 
[K_{1}; K_{2}] = K W_{K}, 
\quad 
\tilde{V} = V W_{V}
\]
Here, $W_{Q}, W_{K}, W_{V} \in \mathbb{R}^{d_{\text{model}} \times 2d}$ are learnable projection matrices, and $\lambda$ is a suppression coefficient that is re-parameterized for numerical stability across layers.

We apply this differential attention mechanism in two complementary directions:

\begin{itemize}
    \item \textbf{Path2Sub}: Pathway embeddings attend to drug substructures, with pathways serving as queries and substructures as keys and values 
    ($Q = \mathbf{H}_{\text{path}},\; K, V = \mathbf{H}_{\text{sub}}$):
    \[
    \mathbf{H}_{P \rightarrow S} = \text{Path2Sub}(\mathbf{H}_{\text{path}}, \mathbf{H}_{\text{sub}})
    \in \mathbb{R}^{N_{p} \times d}
    \]

    \item \textbf{Drug2Path}: Drug-level embeddings attend to pathway representations, with drugs as queries and pathways as keys and values
    \((Q=\mathbf{H}_{\text{drug}},\; K,V=\mathbf{H}_{\text{path}})\):
\[
\mathbf{H}_{D \rightarrow P}
= \text{Drug2Path}(\mathbf{H}_{\text{drug}}, \mathbf{H}_{\text{path}})
\in \mathbb{R}^{1 \times d}
\]
\end{itemize}

Here, $d$ denotes the output dimension of the attention module. By jointly modeling pathway-conditioned substructure relevance and drug-conditioned pathway activation, this dual-view differential cross-attention framework yields robust, biologically contextualized representations that are well suited for downstream drug response prediction.

\subsection{Drug response prediction}

The outputs of the bidirectional differential cross-attention module are aggregated into fixed-length representations for downstream regression. Pathway-, drug-, and substructure-level features are summarized using mean pooling and concatenated into a unified embedding, which is fed into a multi-layer perceptron (MLP) to predict drug response ($\ln(IC_{50})$). This design yields a compact and expressive feature space in which heterogeneous biological signals-pathway context, molecular representation, and chemical substructures-are jointly leveraged for accurate drug response regression.

\subsection{Experimental setup}
\label{subsec:experimental}

\subsubsection{Dataset split}
To evaluate model generalization, we adopted a fixed holdout test set with repeated train-validation splits at the cell line-drug pair level. Specifically, the full dataset was first partitioned into training, validation, and test sets using a 3:1:1 ratio, where the test set was held fixed throughout all experiments and used exclusively for final performance evaluation.
The remaining training-validation portion was then randomly split into training and validation subsets five times, each with a distinct random seed, to account for variability in model optimization. Reported performance metrics represent the mean and standard deviation across these five repeated runs. This protocol ensures fair comparison across models while providing a robust estimate of predictive performance under a consistent test distribution.

\subsubsection{Evaluation metrics}
Model performance was evaluated using standard regression metrics for drug response prediction, including Root Mean Squared Error (RMSE), Pearson's correlation coefficient (PCC), and Spearman's rank correlation coefficient (SCC). RMSE measures numerical prediction accuracy, while PCC and SCC quantify linear and rank-based agreement between predicted and observed responses across cell line-drug pairs. Together, these metrics provide a comprehensive assessment of both prediction accuracy and consistency with experimental measurements. 

\vspace{1em}

\section{Results}\label{sec:results}

\subsection{Performance comparison}

We evaluated the predictive accuracy and generalization capability of DiSPA under four standard data split settings-Random, Cell line-blind, Drug-blind, and Disjoint-set-and compared it against representative baselines, including DeepTTA, DRPreter, DIPK, and DEERS (Table~\ref{tab:performance}). These splits progressively assess model robustness under increasing extrapolation difficulty, ranging from random hold-out to simultaneous exclusion of unseen drugs and cell lines. Detailed descriptions of the baseline models and their modeling characteristics are provided in Approach~\ref{sapp:comparison_methods}.

\begin{figure*}
    \centering
    \includegraphics[
        width=\linewidth,
        keepaspectratio
    ]{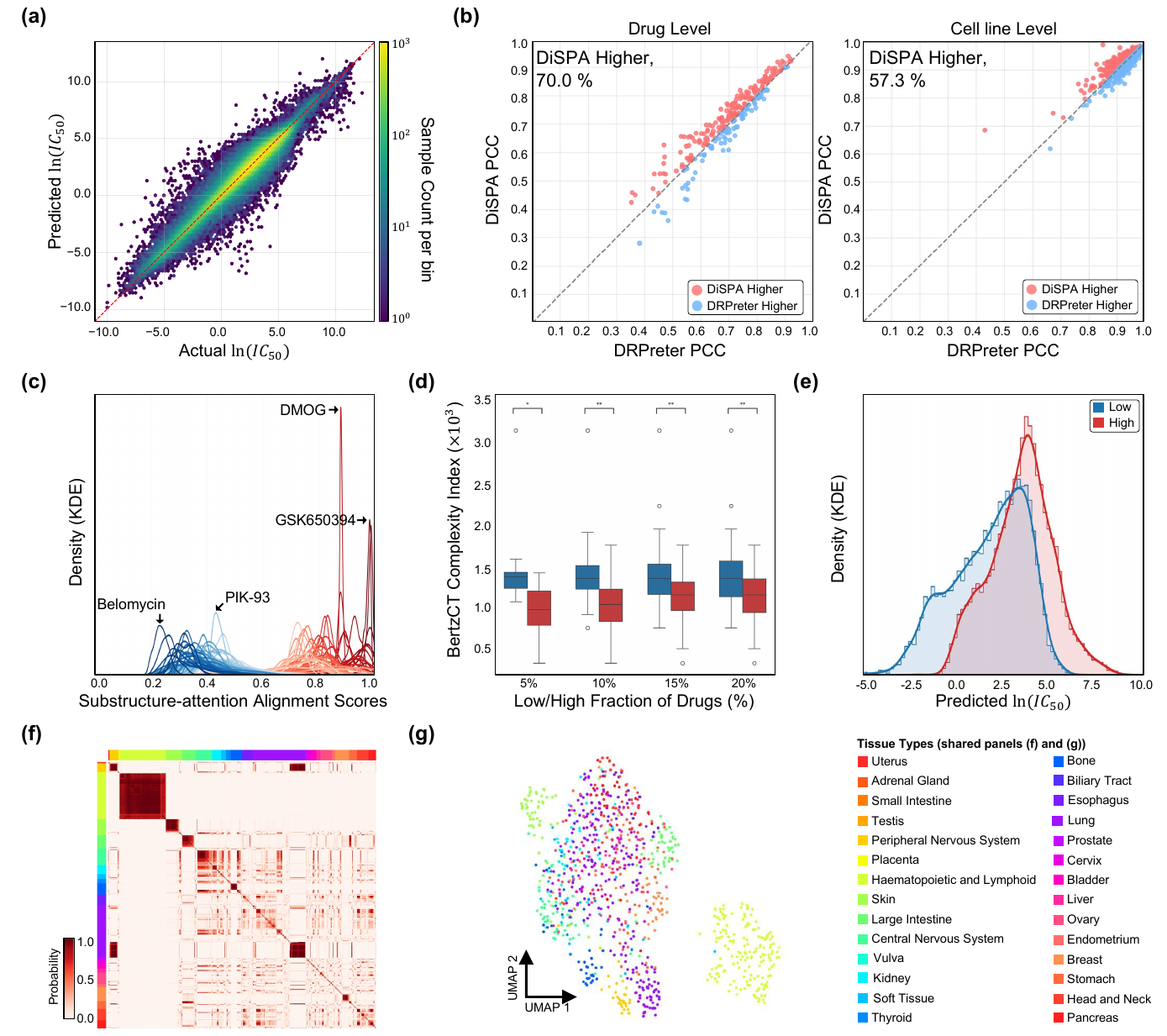}
    \caption{
    {\bf Predictive performance and global organization of drug-cell line interactions learned by DiSPA.}
    (a) Scatter plot comparing predicted and observed $\ln(IC_{50})$ values across all drug-cell line pairs under the Random split, showing strong agreement across the full range.
    (b) Pairwise comparison of PCC between DiSPA and DRPreter at the drug level (left) and cell line level (right), with proportions indicating cases where DiSPA achieves higher correlation.
    (c) Kernel density estimates of substructure-level attention alignment scores across drugs, showing a bimodal distribution with subsets of compounds exhibiting strong attention-similarity correlation (structure-driven) and others showing weaker correlation (context-dependent).
    (d) Comparison of BertzCT structural complexity between drugs stratified by substructure-attention alignment, showing higher complexity in low-alignment compounds.
    (e) Density distributions of predicted $\ln(IC_{50})$ values for representative structure-driven and context-driven drugs across sensitive cell lines, highlighting divergent sensitivity patterns associated with different attention regimes.
    (f) Consensus clustering of cell line attention patterns aggregated across all drugs, revealing a block-diagonal organization corresponding to major tissue lineages.
    (g) UMAP visualization of cell lines based on attention-derived similarity, showing clustering consistent with tissue annotations despite the absence of explicit tissue labels during model training.
    }
    \label{fig:performance}
\end{figure*}

Under the Random split, DiSPA achieved the best overall performance, attaining the lowest prediction error and the highest correlation with measured responses. As shown in Figure~\ref{fig:performance}a, predicted $\ln(IC_{50})$ values closely follow experimental measurements across the full dynamic range, indicating accurate regression behavior. DiSPA also maintained competitive performance in the Cell line-blind setting, demonstrating effective transfer to previously unseen cellular contexts. To further characterize the split settings, we compared scaffold overlap between training and test drugs and gene expression similarity between training and test samples across splits (Figures~\ref{sfig:scaffold_split_analysis} and \ref{sfig:cellline_correlation}).

Performance differences became more pronounced in the Drug-blind and Disjoint-set settings, which require extrapolation to unseen chemical structures. In both scenarios, DiSPA consistently outperformed all baseline methods, achieving the lowest RMSE and the highest correlation metrics among compared models (Table~\ref{tab:performance}). This trend was robust across split seeds, and the broader Drug-blind variance reflected a few low-similarity, high-error drugs (Figure~\ref{sfig:drugblind_variation}, Table~\ref{stab:seed_robustness} and Method \ref{smeth:scaffold_similarity}). To assess whether these gains were uniformly distributed, we further examined performance at the drug and cell line levels. As illustrated in Figure~\ref{fig:performance}b, DiSPA achieved higher PCC than DRPreter for 70.0\% of drugs and 57.3\% of cell lines, indicating that improvements are systematic rather than driven by a small subset of cases.

To evaluate generalization beyond the primary GDSC benchmark, we additionally assessed DiSPA on an independent CTRP \citep{seashore2015harnessing} evaluation and a cross-dataset evaluation on the integrated dataset from CCLE \citep{barretina2012cancer} and PRISM \citep{yu2016high}, where it showed strong overall performance under both external settings (Tables~\ref{stab:ctrp} and~\ref{stab:ccle}).
These results suggest that the model generalizes beyond the primary benchmark, although further validation across additional datasets would be beneficial.

We next investigated whether learned attention patterns reflect meaningful chemical and biological organization. Figure~\ref{fig:performance}c illustrates the relationship between chemical substructure similarity and attention pattern alignment, revealing a distinct separation between ``structure-driven'' compounds and those with more diffuse, context-dependent attention.
This distinction is strongly correlated with structural properties; as shown in Figure~\ref{fig:performance}d, drugs with high substructure-attention alignment exhibit significantly lower BertzCT complexity indices. These differences translate into divergent sensitivity patterns across sensitive cell lines, as captured in the predicted $\ln(IC_{50})$ heatmap (Figure~\ref{fig:performance}e). The complete list of drugs stratified by substructure-attention alignment is summarized in Table~\ref{stab:drug_alignment_list}.

Finally, we investigated the global organization of biological context encoded by the model. Consensus clustering of cell line attention patterns across all drugs reveals a block-diagonal structure corresponding to major tissue lineages (Figure~\ref{fig:performance}f). This organization is further supported by UMAP visualization based on attention-derived similarity, where cell lines cluster according to tissue annotations despite the absence of explicit tissue labels during training (Figure~\ref{fig:performance}g). Collectively, these results indicate that DiSPA not only improves predictive accuracy on the primary benchmark but also learns structured, biologically coherent representations of drug-cell line interactions that generalize across datasets.

\begin{table*}[!t]
  \centering
  \caption{Performance comparison of prediction models under four data split settings (Random, Cell line-blind, Drug-blind, and Disjoint-set), evaluated using RMSE, PCC, SCC.}
  \setlength{\tabcolsep}{8pt}
  \renewcommand{\arraystretch}{1.1}

  \resizebox{1.0\textwidth}{!}{%
  \begin{tabular}{lcccccc}
    \toprule
    & \multicolumn{3}{c}{\textbf{Random}} 
    & \multicolumn{3}{c}{\textbf{Cell line-blind}} \\
    \cmidrule(lr){2-4}\cmidrule(lr){5-7}
    \textbf{Models}
      & \textbf{RMSE (↓)} & \textbf{PCC (↑)} & \textbf{SCC (↑)}
      & \textbf{RMSE (↓)} & \textbf{PCC (↑)} & \textbf{SCC (↑)} \\
    \midrule
    DEERS
      & 1.6130$\pm$0.0101 & 0.8265$\pm$0.0040 & 0.8832$\pm$0.0013
      & 1.7077$\pm$0.0289 & 0.8061$\pm$0.0034 & 0.8330$\pm$0.0036 \\
    DIPK
      & 0.9631$\pm$0.0077 & 0.9315$\pm$0.0011 & 0.9150$\pm$0.0011
      & 1.2865$\pm$0.0091 & 0.8732$\pm$0.0015 & 0.8468$\pm$0.0018 \\
    DeepTTA
      & 0.9528$\pm$0.0026 & 0.9333$\pm$0.0004 & 0.9166$\pm$0.0004
      & \underline{1.2496$\pm$0.0091} & \underline{0.8814$\pm$0.0018} & \underline{0.8548$\pm$0.0016} \\
    DRPreter
      & \underline{0.9295$\pm$0.0054} & \underline{0.9363$\pm$0.0007} & \underline{0.9199$\pm$0.0003}
      & 1.2916$\pm$0.0087 & 0.8722$\pm$0.0016 & 0.8453$\pm$0.0022 \\
    \midrule
    \textbf{DiSPA}
      & \textbf{0.9059$\pm$0.0023} & \textbf{0.9399$\pm$0.0004} & \textbf{0.9236$\pm$0.0003}
      & \textbf{1.2448$\pm$0.0047} & \textbf{0.8821$\pm$0.0009} & \textbf{0.8565$\pm$0.0012} \\
  \end{tabular}}

  \vspace{1mm}

  \resizebox{1.0\textwidth}{!}{%
  \begin{tabular}{lcccccc}
    \toprule
    & \multicolumn{3}{c}{\textbf{Drug-blind}} 
    & \multicolumn{3}{c}{\textbf{Disjoint-set}} \\
    \cmidrule(lr){2-4}\cmidrule(lr){5-7}
    \textbf{Models}
      & \textbf{RMSE (↓)} & \textbf{PCC (↑)} & \textbf{SCC (↑)}
      & \textbf{RMSE (↓)} & \textbf{PCC (↑)} & \textbf{SCC (↑)} \\
    \midrule
    DEERS
      & 2.5490$\pm$0.0294 & 0.3608$\pm$0.0427 & 0.3037$\pm$0.0259
      & 2.5594$\pm$0.0167 & 0.3346$\pm$0.0150 & 0.2444$\pm$0.0204 \\
    DIPK
      & 2.5403$\pm$0.1365 & \underline{0.4431$\pm$0.0768} & \underline{0.3565$\pm$0.0766}
      & \underline{2.5314$\pm$0.0228} & \underline{0.4378$\pm$0.0238} & \underline{0.3247$\pm$0.0169} \\
    DeepTTA
      & \underline{2.5384$\pm$0.1479} & 0.3926$\pm$0.0980 & 0.3086$\pm$0.0775
      & 2.6202$\pm$0.0826 & 0.3575$\pm$0.0550 & 0.2681$\pm$0.0361 \\
    DRPreter
      & 2.5650$\pm$0.0578 & 0.3867$\pm$0.0422 & 0.2894$\pm$0.0649
      & 2.6241$\pm$0.0784 & 0.3840$\pm$0.0349 & 0.2974$\pm$0.0448 \\
    \midrule
    \textbf{DiSPA}
      & \textbf{2.5152$\pm$0.1888} & \textbf{0.4649$\pm$0.0723} & \textbf{0.3673$\pm$0.0839}
      & \textbf{2.4532$\pm$0.0750} & \textbf{0.4776$\pm$0.0468} & \textbf{0.3945$\pm$0.0518} \\
    \bottomrule
  \end{tabular}}
  \label{tab:performance}
\end{table*}

\subsection{Mechanistic insights of substructure-pathway attention}
Quantitative analyses indicated that differential cross-attention produced more concentrated attention and more consistent prioritization of target pathways than vanilla cross-attention in both Path2Sub and Drug2Path modules, providing exploratory evidence of systematic differences in attention patterns without implying direct mechanistic interpretation (Figures ~\ref{sfig:attention_concentration}, \ref{sfig:cumulative_attention_curve} and \ref{sfig:target_pathway_prioritization}).

\begin{figure*}
    \centering
    \includegraphics[width=\linewidth]{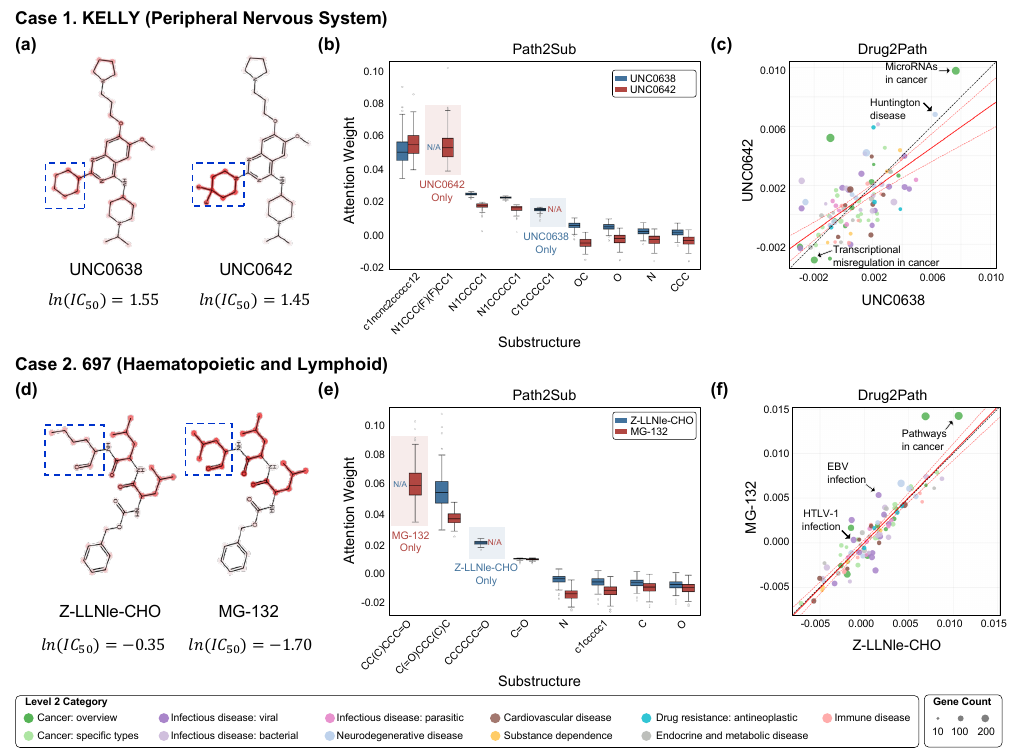}
    \caption{
    {\bf Mechanistic interpretation of substructure-pathway interactions learned by DiSPA.}
    Representative case studies of chemically similar drug pairs with divergent predicted responses.
    \textbf{Case~1} (KELLY, peripheral nervous system): 
    (a) Chemical structures of UNC0638 and UNC0642 with regions highlighted based on Path2Sub attention weights and potential activity-cliffs. 
    (b) Path2Sub attention weights showing differential emphasis on specific substructures. 
    (c) Drug2Path attention comparison illustrating pathway-level consistency and divergence linked to activity differences.
    \textbf{Case~2} (697, haematopoietic and lymphoid): 
    (d) Chemical structures of Z-LLNle-CHO and MG-132 with regions highlighted based on Path2Sub attention weights and potential activity-cliffs.
    (e) Path2Sub attention distributions highlighting substructure-specific differences. 
    (f) Drug2Path attention comparison demonstrating context-dependent pathway-level attention patterns.
    }
    \label{fig:substructure_analysis}
\end{figure*}

To examine whether the differential cross-attention learned by DiSPA reflects biologically meaningful organization, we conducted focused case analyses on representative drug pairs within specific cellular contexts (Figure~\ref{fig:substructure_analysis}). In each case, we selected chemically related compounds associated with a shared biological process but exhibiting differences in predicted activity, allowing assessment of how substructural variation is reflected in pathway-level attention patterns. Additional examples on drug pairs-TGX221/AZD6482 and Imatinib/Masitinib-are provided in Figure~\ref{sfig:additional_case_study}.

\paragraph{Case 1: G9a/GLP inhibitors in neuroblastoma (KELLY).}
UNC0638 and UNC0642 are structurally related G9a/GLP (EHMT2) inhibitors that differ in local ring substitutions influencing cellular permeability and Peripheral Nervous System (PNS) penetration. Figure~\ref{fig:substructure_analysis}a highlights putative activity-cliff-associated regions alongside differences in Path2Sub attention weights in KELLY cells. Path2Sub attention distributions show partially overlapping yet distinct emphasis on specific substructures (Figure~\ref{fig:substructure_analysis}b), while Drug2Path attention comparisons exhibit moderate dispersion around the diagonal (Figure~\ref{fig:substructure_analysis}c), indicating partial agreement in pathway relevance. Attention to neurodegeneration-related pathways, including Huntington disease, is consistent with prior reports linking G9a inhibition to H3K9me2 regulation and neuroprotective transcriptional programs, with UNC0642 exhibiting enhanced CNS activity relative to UNC0638 \citep{bellver2024g9a}.

\paragraph{Case 2: Proteasome inhibitors in haematopoietic cells (697).}
Z-LLNle-CHO and MG-132 are peptide-based proteasome inhibitors with distinct reactive groups and peptide backbones (Figure~\ref{fig:substructure_analysis}d). In 697 cells, Path2Sub attention assigns high weights to substructures associated with proteasome function and protein degradation (Figure~\ref{fig:substructure_analysis}e). Drug2Path attention comparisons show strong diagonal alignment (Figure~\ref{fig:substructure_analysis}f), indicating conserved pathway relevance despite structural differences. Notably, pathways related to lymphoid malignancy and NF-$\kappa$B signaling, including the Human T-cell leukemia virus 1 (HTLV-1) infection pathway, receive consistent attention, in line with prior studies showing that MG-132 suppresses NF-$\kappa$B activity via inhibition of I$\kappa$B$\alpha$ degradation and induces apoptosis in hematological cancers, while Z-LLNle-CHO modulates proteostasis through calpain inhibition \citep{fan2024mg132}.

Across all four cases, DiSPA assigns consistent pathway-level attention to drugs targeting the same biological process, while remaining sensitive to substructural differences that modulate pathway emphasis. Importantly, these patterns emerge without explicit drug-target supervision and correspond closely to established biochemical and cellular mechanisms. Together, Figure~\ref{fig:substructure_analysis} demonstrates that differential substructure-pathway attention provides a coherent and biologically grounded view of how chemical structures and cellular contexts jointly shape drug response.

\subsection{Generalization potential of DiSPA to single-cell and spatial context}\label{subsec:application} 

To evaluate whether DiSPA learns transferable drug-pathway representations, we applied the model-trained exclusively on bulk RNA-seq drug response data-to spatial and single-cell transcriptomics datasets without retraining. This analysis assesses whether pathway-conditioned drug response patterns learned at the bulk level transfer to higher-resolution spatial and cellular contexts. Detailed preprocessing procedures are described in Approach~\ref{sapp:data_preprocessing}.

\begin{figure*}
\centering
\includegraphics[width=\linewidth]{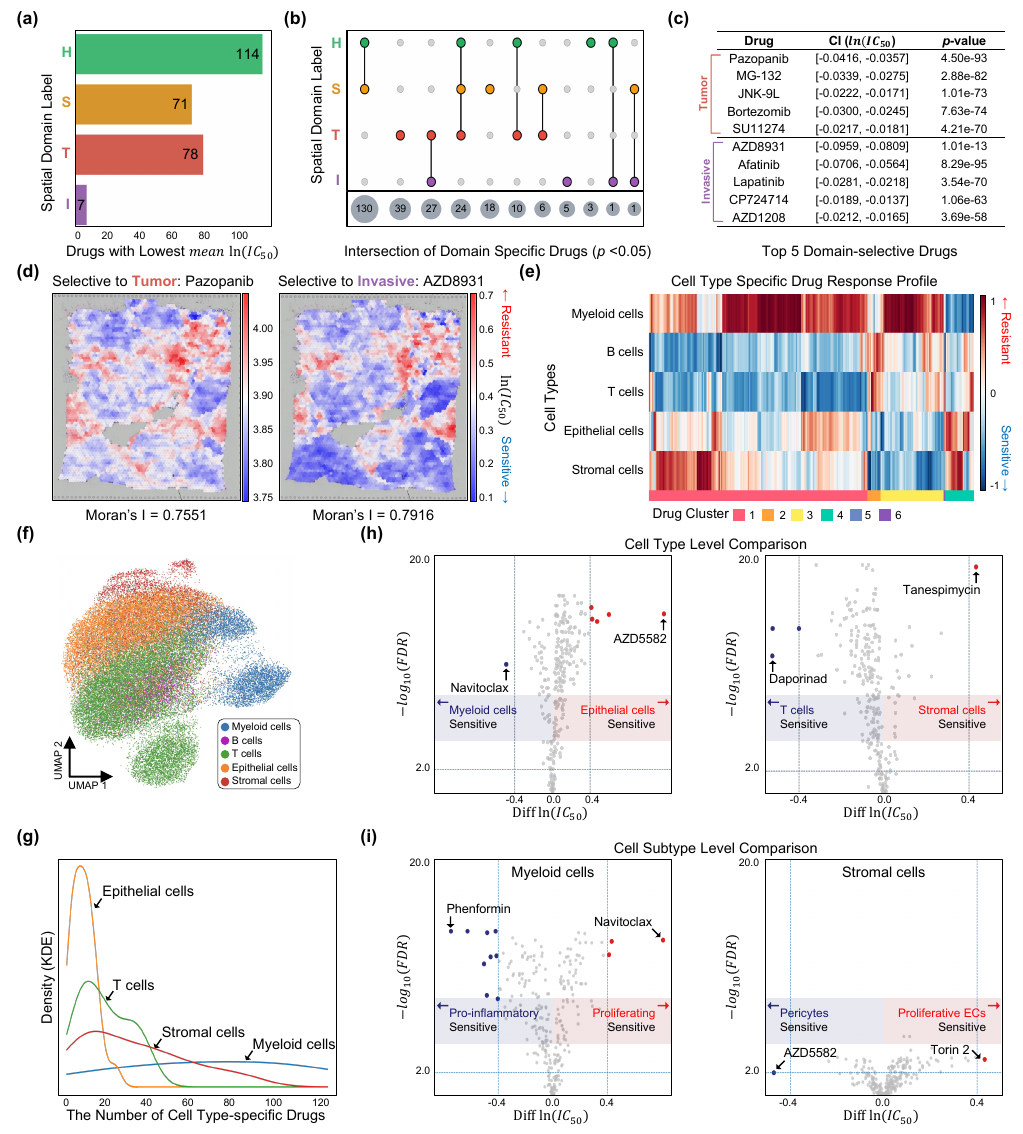}
\caption{
\textbf{Transfer of bulk-trained drug response prediction to spatial and single-cell transcriptomics.}
(a) Counts of spatial domain-selective drugs in an invasive ductal carcinoma spatial transcriptomics dataset.
(b) Overlap of domain-selective drugs ($p < 0.05$).
(c) Top-ranked domain-selective compounds with $\ln(IC_{50})$ differences.
(d) Spatial maps of predicted sensitivity for representative tumor- and invasive-selective drugs.
(e) Cell type-resolved drug response heatmap in a colorectal cancer single-cell RNA-seq atlas.
(f) UMAP of single cells based on predicted drug response profiles.
(g) Distribution of cell type-selective drug counts.
(h, i) Differential drug sensitivity across major cell types and their subtypes.
Predictions are generated using a DiSPA model trained exclusively on bulk RNA-seq data.
}
\label{fig:application}
\end{figure*}

We first applied DiSPA to an invasive ductal carcinoma (IDC) spatial transcriptomics dataset \citep{xun2023reconstruction}. Predicted drug responses exhibited pronounced spatial heterogeneity across annotated tissue domains, including healthy, surrounding tumor, tumor, and invasive regions, as reflected by the number of drugs with the lowest mean predicted $\ln(IC_{50})$ values in each domain (Figure~\ref{fig:application}a). Comparison across domains revealed uneven distributions of sensitive compounds, suggesting domain-specific pharmacological vulnerabilities inferred from spatially resolved expression profiles.

To characterize domain selectivity, we identified drugs exhibiting statistically significant differences in predicted response between domains (one-sided Wilcoxon rank-sum test with FDR correction; $p<0.05$). Intersection analysis revealed both shared and domain-specific sets of selective drugs across spatial compartments (Figure~\ref{fig:application}b), with the top-ranked domain-selective compounds summarized by effect size and statistical significance (Figure~\ref{fig:application}c). Spatial projections of representative tumor- and invasive-selective drugs demonstrated coherent, localized sensitivity patterns aligned with pathological tissue structure, supported by high Moran's I statistics (Figure~\ref{fig:application}d). Detailed spatial response maps for additional domain-selective drugs and the corresponding manual domain annotations are provided in Figures~\ref{sfig:drug_response_IDC} and \ref{sfig:groundtruth_IDC}. Prior literature provides partial support for zero-shot transfer; however, further experimental validation is required (Table~\ref{stab:idc_breast_cancer_evidence}).

We next examined whether DiSPA generalizes to single-cell resolution by applying it to a large colorectal cancer single-cell RNA-seq atlas \citep{lee2020lineage}. Aggregation of predicted responses at the cell type level revealed distinct drug sensitivity landscapes across myeloid cell, T cell, B cell, epithelial, and stromal cell (Figure~\ref{fig:application}e). Cells were embedded using PCA followed by UMAP based on predicted drug response profiles, forming well-separated clusters corresponding to annotated major cell types (Figure~\ref{fig:application}f), despite the absence of cell type supervision during model training. 
Cell type-specific projections of the shared UMAP embedding are provided in Figure~\ref{sfig:cell_umap}, with additional subtype-level projections illustrating intra-cell type heterogeneity shown in Figure~\ref{sfig:cell_sub_umap}.

The distribution of selectively sensitive drugs varied across cell types, with epithelial and stromal populations exhibiting broader drug-specific responses compared to immune compartments (Figure~\ref{fig:application}g). Differential drug sensitivity analyses between selected major cell type pairs identified compounds with preferential predicted efficacy, as illustrated by volcano plots comparing epithelial versus myeloid cells and stromal versus T cells (Figure~\ref{fig:application}h). Extending this analysis to finer cellular resolution further revealed subtype-level differences within myeloid and stromal compartments (Figure~\ref{fig:application}i), indicating that predicted pharmacological patterns are preserved across hierarchical cellular organization.

Collectively, these results demonstrate that DiSPA transfers bulk-trained drug response representations to spatial and single-cell transcriptomics settings, enabling systematic analysis of region- and cell type-specific pharmacological sensitivity without retraining.

\section{Discussion}

DiSPA addresses a central limitation of existing drug response prediction models: the lack of explicit modeling of interactions between chemical structures and biological context. While prior approaches integrate chemical and transcriptomic features, they often rely on static drug representations or late-fusion strategies, which can constrain both generalization and interpretability. By introducing a dual-view differential cross-attention framework, DiSPA jointly conditions chemical substructures on pathway-level gene expression and pathway representations on drug context, enabling integrated modeling of chemical and biological information during representation learning.

Across multiple evaluation settings, DiSPA shows competitive predictive performance, with the largest gains observed in Drug-blind and Disjoint-set splits that require extrapolation to unseen drugs and cell lines. These results suggest that pathway-conditioned chemical representations contribute to robustness under distribution shift. Controlled comparisons and ablations supported the benefit of differential bidirectional attention over simpler interaction baselines and vanilla cross-attention (Tables~\ref{stab:interaction_modules}, ~\ref{stab:attention_ablation}, and ~\ref{stab:ablation}).

Beyond accuracy, the learned attention patterns exhibit structured organization at both chemical and biological levels. Substructure-attention alignment varies across compounds, with weaker alignment observed for structurally complex molecules, indicating systematic dependence on molecular properties. At the biological level, pathway-conditioned attention patterns group cell lines in a manner consistent with tissue annotations, despite the absence of explicit tissue supervision. These observations suggest that the model learns representations partially consistent with known biological organization, although they do not imply direct mechanistic causality.

DiSPA also generalizes to higher-resolution transcriptomic modalities. When applied without retraining to spatial transcriptomics and single-cell RNA-seq data, the model produces structured region- and cell type-specific drug response patterns. While exploratory, these results indicate that representations learned from bulk RNA-seq can transfer to more heterogeneous cellular contexts, supporting the use of DiSPA for hypothesis generation across spatial and cellular scales.

Several limitations should be noted. First, training and primary evaluation rely on a single pharmacogenomic resource, and broader cross-dataset validation remains an important direction for future work. Second, attention weights reflect model-derived attribution rather than direct evidence of molecular interactions, requiring experimental validation for mechanistic interpretation. Finally, transfer analyses to spatial and single-cell data lack ground-truth drug sensitivity measurements and should therefore be interpreted qualitatively.

\section{Conclusion}
We presented DiSPA, a differential cross-attention framework for integrating pathway-level gene expression and substructural drug representations in drug response prediction. The model achieves strong predictive performance and learns coherent pathway-substructure attention patterns without relying on explicit drug-target information. These representations enable structured analysis of drug response with respect to both chemical and pathway-level biological context. Overall, DiSPA provides a unified framework for learning structured representations across chemical and biological modalities, and establishes a principled basis for future studies on cross-dataset generalization, multimodal integration, and systematic evaluation of attention-based pharmacogenomic models.


\section{Competing interests}
No competing interest is declared.

\section{Author contributions statement}
Y.H. and S.Lim conceived the study. 
Y.H., S.K., and S.Y. performed the experiments. 
Y.H., S.Lee, and S.Lim analyzed the data. 
Y.H., E.J., and S.Lim drafted the manuscript. 
S.Lim supervised the project and secured funding. 
All authors reviewed and approved the final manuscript.

\section{Acknowledgments}
This work was supported by the National Research Foundation of Korea (NRF) grants funded by the Korean government (MSIT) (RS-2025-00560523, RS-2025-18732993, and RS-2022-NR067309). This work was also supported by the Institute of Information \& Communications Technology Planning \& Evaluation (IITP) grants funded by the Korean government (MSIT) under the Artificial Intelligence Convergence Innovation Human Resources Development program (IITP-2025-RS-2023-00254592) and the Information Technology Research Center (ITRC) program (IITP-2025-RS-2020-11201789).

\beginsupplement
\refstepcounter{approach}
\label{sapp:data_availability}

\refstepcounter{approach}
\label{sapp:data_preprocessing}

\refstepcounter{approach}
\label{sapp:comparison_methods}

\refstepcounter{figure}
\label{sfig:scaffold_split_analysis}

\refstepcounter{figure}
\label{sfig:cellline_correlation}

\refstepcounter{figure}
\label{sfig:drugblind_variation}

\refstepcounter{figure}
\label{sfig:attention_concentration}

\refstepcounter{figure}
\label{sfig:cumulative_attention_curve}

\refstepcounter{figure}
\label{sfig:target_pathway_prioritization}

\refstepcounter{figure}
\label{sfig:additional_case_study}

\refstepcounter{figure}
\label{sfig:drug_response_IDC}

\refstepcounter{figure}
\label{sfig:groundtruth_IDC}

\refstepcounter{figure}
\label{sfig:cell_umap}

\refstepcounter{figure}
\label{sfig:cell_sub_umap}

\refstepcounter{table}
\label{stab:pathway_gene_mapping}

\refstepcounter{table}
\label{stab:kegg_category_performance}

\refstepcounter{table}
\label{stab:kegg_full_vs_cat6}

\refstepcounter{table}
\label{stab:brics_excluded_drugs}

\refstepcounter{table}
\label{stab:seed_robustness}

\refstepcounter{table}
\label{stab:ctrp}

\refstepcounter{table}
\label{stab:ccle}

\refstepcounter{table}
\label{stab:drug_alignment_list}

\refstepcounter{table}
\label{stab:idc_breast_cancer_evidence}

\refstepcounter{table}
\label{stab:interaction_modules}

\refstepcounter{table}
\label{stab:attention_ablation}

\refstepcounter{table}
\label{stab:ablation}

\refstepcounter{method} 
\label{smeth:scaffold_similarity}

\bibliographystyle{abbrvnat}
\bibliography{DiSPA_ref}

\end{document}